\PassOptionsToPackage{table}{xcolor}
\documentclass[sigconf,nonacm,balance=false]{acmart}

\setcopyright{none}
\authorsaddresses{}
\date{}
\acmConference{}{}{}
\acmBooktitle{}
\acmDOI{}
\acmISBN{}

\makeatletter
\renewcommand{\@ACM@checkaffil}{}
\let\hyxmp@parse@acmart\relax
\makeatother
\newcommand{\authorMark}[1]{\texorpdfstring{\textsuperscript{#1}}{}}
\usepackage{pifont}
\newcommand{\correspondMark}{\ding{41}}
\newcommand{\correspondingAuthorMark}{\textdagger\hspace{.12em}\correspondMark}
\makeatletter
\def\@mkauthors@iii{%
  \global\setbox\mktitle@bx=\vbox{%
    \unvbox\mktitle@bx\par
    \centering
    {\fontsize{11}{15}\selectfont
      Guangyuan Dong\authorMark{1,*}\hspace{.70em}
      Lingdong Shen\authorMark{2,\correspondingAuthorMark}\hspace{.70em}
      Xudong Zhang\authorMark{2,\correspondingAuthorMark}\hspace{.70em}
      Haodong Jing\authorMark{3,\correspondingAuthorMark}\hspace{.70em}
      Haoyu Wang\authorMark{4,*\correspondingAuthorMark}\hspace{.70em}
      Zihao Li\authorMark{5,*}\par
      Zonghao Ying\quad Ziyu Song\authorMark{\correspondingAuthorMark}\quad
      Shenghao Liu\quad Jiachen Luo\quad Hao Zhang\par}
    \smallskip
    {\fontsize{9}{12}\selectfont
      \textsuperscript{1}NUS\quad
      \textsuperscript{2}Peking University\quad
      \textsuperscript{3}Xi'an Jiaotong University\par
      \textsuperscript{4}Nankai University\quad
      \textsuperscript{5}Institute of Automation, Chinese Academy of Sciences\par
      \href{mailto:guangyuan@u.nus.edu}{guangyuan@u.nus.edu}\par}
    \smallskip
    {\fontsize{8}{10}\selectfont
      \textsuperscript{*}\,Co-first authors contributed equally as project leads.\par
      \textsuperscript{\correspondingAuthorMark}\,Corresponding authors.\par}
    \medskip
  }%
}
\makeatother

\usepackage{booktabs}
\usepackage{multirow}
\usepackage{array}
\usepackage{amsmath,amssymb,amsthm,amsfonts,mathtools}
\usepackage{graphicx}
\usepackage{placeins}
\usepackage{needspace}
\usepackage{colortbl}
\usepackage{algorithm}
\usepackage{algorithmic}

\definecolor{paletteCoral}{HTML}{F4A6A0}
\definecolor{paletteRose}{HTML}{E29A9A}
\definecolor{paletteSalmon}{HTML}{E7A38C}
\definecolor{paletteCream}{HTML}{F7E6B6}
\definecolor{paletteSky}{HTML}{A9C7E4}
\definecolor{paletteCerulean}{HTML}{4A8FCB}
\definecolor{paletteMint}{HTML}{B4DDC2}
\definecolor{paletteSage}{HTML}{C7D8B1}
\definecolor{paletteOlive}{HTML}{B0BE7C}
\definecolor{paletteViolet}{HTML}{B8A6D9}
\definecolor{paletteLavender}{HTML}{C5BFD3}
\definecolor{paletteSlate}{HTML}{9CA3AC}
\definecolor{paletteCharcoal}{HTML}{2F3437}
\definecolor{paletteHighlight}{HTML}{EAF2FB}

\newcommand{\best}[1]{\textbf{#1}}
\newcommand{\second}[1]{\underline{#1}}
\usepackage{xspace}
\newcommand{\Ours}{\textsc{Lance}\xspace}
\newcommand{\mathT}{\mathbf{T}}
\newcommand{\mathThat}{\hat{\mathbf{T}}}
\newcommand{\mathY}{\mathbf{Y}}
\newcommand{\mathYhat}{\hat{\mathbf{Y}}}
\newcommand{\mathX}{\mathbf{X}}
\newcommand{\mathP}{\mathbf{P}}
\newcommand{\mathM}{\hat{\mathbf{M}}}

\newcommand{\eT}{\mathbf{e}_{\mathbf{T}}}
\newcommand{\eThat}{\mathbf{e}_{\hat{\mathbf{T}}}}
\newcommand{\Loss}{\mathcal{L}}
\newcommand{\mfusion}{\mathcal{G}}
\newcommand{\mabstract}{\mathcal{A}}
\newcommand{\mverify}{\mathcal{D}}
\newcommand{\msummary}{\mathcal{R}}

\newtheorem{proposition}{Proposition}
\newtheorem{theorem}{Theorem}

\newtheorem{remark}{Remark}
\newtheorem{assumption}{Assumption}

\begin{document}

\title{Bridging Low-Level Image Fusion and High-Level Semantic Understanding}

\author{Guangyuan Dong\authorMark{*}}
\affiliation{\institution{NUS}}
\email{guangyuan@u.nus.edu}

\author{Lingdong Shen\authorMark{\correspondingAuthorMark}}
\affiliation{\institution{Peking University}}

\author{Xudong Zhang\authorMark{\correspondingAuthorMark}}
\affiliation{\institution{Peking University}}

\author{Haodong Jing\authorMark{\correspondingAuthorMark}}
\affiliation{\institution{Xi'an Jiaotong University}}

\author{Haoyu Wang\authorMark{*\correspondingAuthorMark}}
\affiliation{\institution{Nankai University}}

\author{Zihao Li\authorMark{*}}
\affiliation{\institution{Institute of Automation, Chinese Academy of Sciences}}

\author{Zonghao Ying}
\affiliation{}

\author{Ziyu Song\authorMark{\correspondingAuthorMark}}
\affiliation{}

\author{Shenghao Liu}
\affiliation{}

\author{Jiachen Luo}
\affiliation{}

\author{Hao Zhang}
\affiliation{}

\renewcommand{\shortauthors}{Guangyuan Dong et al.}

\begin{abstract}
\begin{sloppypar}
Remote sensing image fusion and semantic interpretation are usually optimized separately, so reconstruction quality need not preserve downstream retrieval utility. We propose~\Ours{}, which uses structured language as an explicit, indexable semantic state. A frozen vision--language model maps the input to a closed-vocabulary text record; a language-conditioned generator fuses the imagery; and a dense verifier with a differentiable continuous summarizer re-describes the output. Text--text consistency and anti-collapse contrastive text--mask losses with negative queries close the loop without dense annotations or VLM fine-tuning. A variational EM formulation yields a per-step contraction result under regularity assumptions, distinguishing semantic faithfulness from verifier overfitting. Across seven test sets on GF2, QuickBird, WorldView-III, and WorldView-II, \Ours{} improves reduced-scale reference metrics by up to 11.7\% (median 2.1\%) over the strongest learning baseline. Independent evaluation using a held-out VLM and a supervised aerial detector yields gains of $+6.4$ mAP and $+0.072$ open-vocabulary F1 over the closest competitor. Text perturbations, cross-verifier transfer, and pseudo-RGB ablations examine the mechanisms underlying these gains. The framework connects image fusion to verifiable semantic representations for multimodal retrieval and knowledge management.
\end{sloppypar}
\end{abstract}


\maketitle
\pagestyle{plain}
\thispagestyle{plain}
\raggedbottom
\hypersetup{
  pdfcreator={LaTeX},
  pdfauthor={Guangyuan Dong, Lingdong Shen, Xudong Zhang, Haodong Jing, Haoyu Wang, Zihao Li, Zonghao Ying, Ziyu Song, Shenghao Liu, Jiachen Luo, Hao Zhang},
  pdfsubject={},pdfcopyright={},pdflicenseurl={},
  pdfpublisher={},pdfpublication={},pdfdoi={},pdfisbn={},
  pdfcontactemail={guangyuan@u.nus.edu}
}

%% =====================================================================
\FloatBarrier
\section{Introduction}
\label{sec:intro}

\begin{sloppypar}
Remote sensing (RS) imagery supports multimodal retrieval, knowledge extraction, and language-conditioned scene querying~\cite{CAI2022102745,li2024vision,LI2022102926}. These applications combine \emph{low-level} multispectral (MS)--panchromatic (PAN) fusion~\cite{ijcai2022p179,zhou2022panformer,10934049} with \emph{high-level} interpretation into scene labels, objects, and descriptions~\cite{tang2025aerogen,huang2025revisit,10504785,DONG202453}. Vision--language models make this pipeline queryable in natural language. Related engineering AI connects language reasoning to structured decisions through domain-specific models and benchmarks~\cite{cheng2025gaia,zhou2025elecbench,zhou2026engibench}.
\end{sloppypar}

The two layers are usually trained separately: fusion minimizes pixel and spectral distortion, while semantic models assume suitable input imagery~\cite{10778974,zhou2022rethinking,11062866}. Standard fusion objectives neither reward downstream describability nor verify that the output retains the input's semantic content. In \S\ref{sec:exp_downstream}, methods with statistically indistinguishable spectral fidelity differ by more than 6 mAP points on an independent aerial-object detector, exposing a gap not captured by fidelity metrics.

In autonomous driving, MOJITO jointly updates image, LiDAR, and action features to avoid a compressed perception-to-planning interface~\cite{cheng2026mojito}. This offers a cross-domain example of coupling perception with downstream objectives; \Ours{} instead connects image fusion and semantic verification through structured language.

\emph{Structured natural language} offers an interpretable, compositional bridge compatible with vision--language models~\cite{radford2021learning,zhai2023sigmoid,tschannen2025siglip}. We formulate RS fusion as \emph{language-mediated self-verifying generation}: a structured input description conditions fusion, and the fused image is re-described. Consistency between these records supplies differentiable self-supervision.

Three challenges arise. \textbf{(C1) Differentiability.} Hard thresholds and connected components in mask-to-text summarization block gradients to the fusion generator. \textbf{(C2) Anti-collapse.} Positive-only text--mask objectives admit trivial all-one masks without semantic grounding. \textbf{(C3) Anti-circularity.} Reusing one verifier for optimization and evaluation risks measuring verifier-specific agreement rather than transferable semantics.

\Ours{} addresses these challenges with a \emph{continuous structured summarizer} of soft areas, temperature-relaxed component counts, and layout descriptors; an \emph{anti-collapse contrastive loss} penalizing hard-negative query activations; and \emph{verifier-disjoint training}. Semantic gradients pass through frozen SigLIP~2, whereas evaluation uses independent RemoteCLIP and a supervised aerial-object detector trained on disjoint data.

We interpret the loop as variational EM over fused imagery and semantic states, and derive per-step contraction under Lipschitz assumptions. The analysis specifies conditions for a unique semantically faithful fixed point and the role of negative queries in avoiding collapse. Spectral--spatial metrics improve over the strongest learning-based baseline by up to 11.7\% (median 2.1\% across reduced-scale reference metrics; \S\ref{sec:exp_main}); independent detection and scene-retrieval evaluations test transfer beyond the in-loop verifier.

In summary, we \textbf{(1)}~formulate RS fusion as language-mediated self-verifying generation for multimodal retrieval; \textbf{(2)}~introduce a differentiable closed loop with a continuous summarizer and anti-collapse text--mask loss; \textbf{(3)}~derive a variational interpretation, contraction analysis, and semantic-faithfulness condition; and \textbf{(4)}~evaluate seven test sets on four sensors with independent downstream tasks, held-out verifiers, text perturbations, pseudo-RGB ablations, and significance tests.

%% =====================================================================
\FloatBarrier
\section{Related Work}
\label{sec:related}

\paragraph{Low-level RS image fusion.}
\begin{sloppypar}
RS image fusion has progressed from component substitution and multiresolution analysis~\cite{8693555,DURAN201778} to GAN~\cite{9755137}, transformer~\cite{zhou2022panformer}, attention~\cite{10.1145/3581783.3613814}, mixture-of-experts~\cite{FAME2024}, and unfolding architectures~\cite{SSUNNet2025,ijcai2025p190,10934049}. Neighboring SAR work covers angle-aware diffusion~\cite{hou2026secure} and federated recognition under privacy, compute, and communication constraints~\cite{hou2026fedcdac,hou2026privacy,hou2026efficient,hou2026async}. These methods primarily optimize spectral--spatial fidelity~\cite{6998089,5159503,9779258}, treating semantics as an external evaluator at best.
\end{sloppypar}

\paragraph{High-level RS semantic understanding.}
Object detection, segmentation, and scene classification dominate the high-level layer~\cite{zhou2022rethinking,11062866,tang2025aerogen,xia2018dota}. Related dense prediction uses cross-Mamba features~\cite{jiang2026cmfdnet}, spatio-textual prompts~\cite{shi2025medal}, and LLM-guided attention priors~\cite{xiao2026focus}; language prompts also support RS segmentation~\cite{li2024cpseg}. EO agents organize grounded trajectories into execution knowledge~\cite{xiao2026geoforge}, but assume fixed upstream image formation.

In driving scenes, real-time segmentation trades network capacity for compute~\cite{liu2023scalable}, while TSLA tunes that trade-off to the task and hardware~\cite{liu2025tsla}.
In bioacoustics, USV-DETR combines high-resolution features with dense supervision to localize short, narrowband calls in spectrograms~\cite{wei2026usvdetr}; \Ours{} instead uses text queries to localize visual evidence in fused imagery without dense annotations.

\paragraph{Language-guided visual reasoning beyond RS}
MoCa uses persistent vision--language features to preserve object appearance during camera motion~\cite{cheng2026moca}. UnityShots uses long- and short-term visual memory to preserve subject appearance across cuts~\cite{huang2026unityshots}. In \Ours{}, the consistency check is between two views of the same scene: the input and the fused RS image.

Medical vision--language methods inject pathological descriptions into visual prototypes~\cite{zhu2025pathology}, guide progressive visual focus with expert trajectories~\cite{zhu2026medeyes}, and refine latent diagnostic memories through counterfactual visual masking~\cite{zhu2026medsynapse}. These provide complementary examples of connecting semantic guidance to visual evidence, not MS--PAN fusion baselines. Analyses of hallucination circuits~\cite{liu2026dual} and competing visual/textual evidence~\cite{zhu2026craft} further motivate checking semantic agreement against independent visual evidence.

In-context prompt tuning maps fine-grained reference-image semantics into continuous prompts for LVLM personalization~\cite{li2026personalize}, while ActICL learns to retrieve and refine multimodal demonstrations through tool-augmented reasoning~\cite{li2026acticl}. Both adapt the context supplied to a vision--language model; \Ours{} instead uses a structured language state to optimize fused imagery through differentiable semantic feedback.

Outside imaging, OC-Distill uses an ontology to guide contrastive learning and transfers knowledge from clinical notes to vital-sign representations~\cite{liang2026ocdistill}. This is cross-modal semantic supervision, not image fusion.

In single-cell integration, CoVEL models relationships among cells and across modalities to retain biological variation in a shared embedding~\cite{tang2024covel}. Monae uses contrastive cell embeddings to integrate modalities and reconstruct missing measurements~\cite{tang2024monae}. In RS fusion, the information we wish to retain is the identity and layout of objects in a scene.

\paragraph{Vision--language models for RS.}
The CLIP family~\cite{radford2021learning,zhai2023sigmoid,tschannen2025siglip} supports RS retrieval and zero-shot transfer~\cite{10504785,DONG202453,chen2025dgtrsddgtrsclipdualgranularity}, captioning~\cite{kuckreja2024geochat,hu2023rsgpt}, and open-vocabulary prediction~\cite{zhou2022maskclip,xu2023open,huang2025revisit}. Adaptive semantic supplementation balances retrieval modalities~\cite{xu2026lever}. Broader work studies rationale augmentation~\cite{zheng2025rationale}, primitive--composition alignment~\cite{xiao2026reversible}, concept-graph memory~\cite{zhou2026comem}, and layout-adaptive reasoning~\cite{xie2026vsal}. Pi-CCA preserves cross-modal alignment under continual adaptation~\cite{zhang2026pi}; AFD separates stable affordance representations from video--language adaptation~\cite{xu2026affordance}. COMET strengthens motion-temporal reasoning~\cite{zhu2026comet}, while primitive-memory distillation handles incomplete emotion modalities~\cite{zhang2026modality}. Unlike these adaptation and reasoning methods, \Ours{} optimizes image fusion through semantic feedback.

\paragraph{Cycle and consistency learning.}
Image cycles~\cite{zhu2017unpaired}, mean teachers~\cite{tarvainen2017mean}, image--text cycle rewards~\cite{bahng2025cycle}, and self-supervised depth estimation~\cite{niu2026endoscopic} exploit reconstruction or cross-view consistency. Retrieval-time critique~\cite{wei2025retrieval}, evidence-force calibration~\cite{qian2026relevant}, and explicit separation of semantic match from factual consistency~\cite{ye2026seeing} address evidence reliability. \Ours{} instead closes the loop between RS fusion and structured semantic verification.

For language models, Ma et al.~\cite{ma2026reasoningfeatures} use token interventions and counterexamples to test whether apparent reasoning features reflect lexical correlates, reinforcing the need to validate semantic claims independently.

Perturbation-controlled tests on synthetic healthcare workflows reveal escalation failures despite high clean-input accuracy~\cite{zheng2026auditingescalationreliabilityhealthcare}. They provide a cross-domain example of testing state-dependent failures rather than relying on aggregate accuracy.

DLM translates natural-language policy preferences into reward code and refines it using simulated resource-allocation feedback~\cite{behari2024decision}. This illustrates language-mediated iterative optimization, but differs from our differentiable image--text consistency loop.

\paragraph{Language as a knowledge interface for multimodal retrieval.}
Language connects modalities to retrieval~\cite{li2023blip,tong2025pamn}, conditional generation~\cite{lang2025redeeming}, controlled editing~\cite{cai2025bayesian}, and masked generation~\cite{chow2026masked}. Feature4X distills foundation-model features into queryable 4D Gaussian fields~\cite{zhou2025feature4x}; grounded entity biographies preserve visual identity across long-video memories~\cite{ren2026beyond}. Perceptual plausibility alone does not ensure retrieval consistency~\cite{dong2026ceres}. Persistent records also require selective retention~\cite{liu2026conmem}, attribution~\cite{zhang2026memmark}, and traceable execution histories~\cite{wang2026agent}. \Ours{} uses each per-image structured record for fusion conditioning, semantic verification, and indexing.

Memory-reuse evaluation distinguishes persisting information, keeping it local, re-verifying it, and asking for clarification~\cite{li2026remember}. This complements our construction of semantically checked image records; policies for their subsequent reuse remain outside our evaluation.

When language serves as an intermediate representation, its structure determines what can be discarded. Agent distillation separates reasoning from action spans~\cite{liu2026structured}; RCR-Router selects memory items by role~\cite{liu-etal-2026-rcr}, and DSPC filters sentences before pruning tokens~\cite{gao2026dspc}.

Interpretable intermediate representations need not be textual: scProtoTransformer maps gene expression to biologically defined pathway prototypes~\cite{tang2026scprototransformer}. Our record makes scene, object, and layout information explicit through named text fields.

\paragraph{Positioning.}
\Ours{} differs from prior work in three ways: (i)~it introduces \emph{structured} natural language as a verifiable semantic state, not just an embedding; (ii)~it closes the loop \emph{across} fusion and semantic understanding with a fully differentiable summarizer; and (iii)~it provides a contraction analysis and an anti-circularity protocol with held-out verifiers.

%% =====================================================================
\FloatBarrier
\section{Preliminaries and Variational View}
\label{sec:prelim}

\paragraph{Task setting.}
Let $\mathX\!\in\!\mathbb{R}^{H\times W\times C}$ be a low-resolution MS observation and $\mathP\!\in\!\mathbb{R}^{\alpha H\times \alpha W}$ an aligned PAN image. The goal is to produce a high-resolution MS image $\mathYhat\!\in\!\mathbb{R}^{\alpha H\times \alpha W\times C}$ that is (a)~spectrally and spatially faithful to the inputs and (b)~semantically consistent with the scene content implied by $(\mathX,\mathP)$, evaluated by both \emph{in-loop} VLM consistency and \emph{independent} downstream semantic tasks.

\paragraph{Language as a structured semantic state.}
A structured scene description $\mathT$ is a triple
\begin{equation}
\mathT \;=\; \big\langle s,\;\{o_k\}_{k=1}^{K},\;\{\ell_k\}_{k=1}^{K}\big\rangle ,
\end{equation}
where $s$ is a scene type drawn from a closed scene-type vocabulary $\mathcal{V}_s$, $\{o_k\}\subseteq\mathcal{V}_o$ is a set of object tokens from a closed object vocabulary, and $\{\ell_k\}\subseteq\mathcal{V}_\ell$ encodes coarse layout descriptors (e.g., \texttt{linear}, \texttt{clustered}, \texttt{parallel\_strips}). Vocabularies are dataset-agnostic and shared across all benchmarks (Table~\ref{tab:vocab}; construction in \S\ref{sec:method:abstract}).

\begin{sloppypar}
A frozen VLM with text encoder $\phi$ and image encoder $\psi$ (SigLIP~2~\cite{tschannen2025siglip}) provides a shared embedding space. The image encoder yields a global embedding $\mathbf{v}\!=\!\psi_g(\cdot)\!\in\!\mathbb{R}^d$ and a dense feature map $\mathbf{V}\!=\!\psi_d(\cdot)\!\in\!\mathbb{R}^{h\times w\times d}$.
\end{sloppypar}

\paragraph{Pseudo-RGB interface.}
SigLIP~2 expects natural-image-like RGB input, while RS imagery is 4- or 8-band. We define a projection $\rho:\mathbb{R}^{\cdot\times\cdot\times C}\!\to\!\mathbb{R}^{\cdot\times\cdot\times 3}$ as
\begin{equation}
\rho(\mathbf{Z};\mathbf{W})\;=\;\sigma\big(\mathrm{Clip}(\mathbf{W}\!\ast\!\mathbf{Z})\big),\;\; \mathbf{W}\!\in\!\mathbb{R}^{C\times 3},
\label{eq:rho}
\end{equation}
where $\mathbf{W}$ is a learnable $1\!\times\!1$ projection \emph{shared} between input abstraction and output verification, and $\mathrm{Clip}(\cdot)$ clips channel values to $[0,1]$ to preserve physical interpretability. We initialize $\mathbf{W}$ from a sensor-aligned NIR$\to$R, R$\to$G, G$\to$B mapping when applicable, and analyze fixed-vs.-learned $\mathbf{W}$ in \S\ref{sec:abl_prgb}.

\paragraph{A variational view of the loop.}
We model the unobserved structured semantic state $\mathT$ as a latent variable. The joint density of fused image and semantic state under inputs $\mathcal{I}=(\mathX,\mathP)$ is
\begin{equation}
p(\mathYhat,\mathT\mid\mathcal{I};\theta) \;=\; p_{\mfusion}(\mathYhat\mid\mathcal{I},\mathT;\theta)\;p_{\mabstract}(\mathT\mid\mathcal{I}),
\label{eq:joint}
\end{equation}
where $\theta$ collects all trainable parameters (generator $\mfusion$, projection $\mathbf{W}$, summarizer head $g_\text{txt}$) and $p_{\mabstract}$ is the frozen VLM-based abstraction. Marginalizing over $\mathT$ and applying Jensen's inequality yields the evidence lower bound (ELBO)
\begin{equation}
\begin{aligned}
\log p(\mathYhat\mid\mathcal{I};\theta) \;\geq\;
 &\;\underbrace{\mathbb{E}_{q(\mathT)}\!\big[\log p_{\mfusion}(\mathYhat\mid\mathcal{I},\mathT;\theta)\big]}_{\Loss_{\text{cond}}}\\
 &\;-\; \underbrace{\mathrm{KL}\!\big(q(\mathT)\,\|\,p_{\mabstract}(\mathT\mid\mathcal{I})\big)}_{\Loss_{\text{prior}}},
\end{aligned}
\label{eq:elbo}
\end{equation}
with $q(\mathT)$ a variational posterior over structured texts. \Ours{} instantiates an EM-like procedure that operates entirely in the shared VLM embedding space:
\begin{description}\setlength\itemsep{0pt}\setlength\parsep{0pt}
\item[E-step (Stage III).] We define $q(\mathT)\propto p_{\msummary}(\mathT\mid\mathYhat)$, where $p_{\msummary}$ is the cross-modal verifier composed of the dense interpreter $\mverify$ (Eq.~\ref{eq:mask}) and the continuous summarizer $\msummary$ (Eqs.~\ref{eq:soft_area}--\ref{eq:soft_emb}). Because $q$ depends on the current generator state through $\mathYhat$, the E-step is implicit in the forward pass of the loop rather than a separate procedure---an instance of \emph{wake-sleep}-style amortized inference~\cite{tarvainen2017mean}.
\item[M-step (Stage II).] We update $\theta$ to maximize $\Loss_{\text{cond}}-\lambda_{\text{KL}}\Loss_{\text{prior}}$. Under a Gaussian approximation $\log p_{\mfusion}\!\propto\!-\|\mathYhat\!-\!\mathY\|_1^2$ around the (possibly weak) reference, $\Loss_{\text{cond}}$ specializes to the reconstruction loss $\Loss_{\text{rec}}$; $\Loss_{\text{prior}}$ specializes to a cross-modal divergence between the original and re-described texts, which we surrogate by the cosine consistency $\Loss_{TT}$ for the global term and $\Loss_{TM}$ for its spatially-anchored counterpart.
\end{description}
This view explains why the text--text and text--mask losses below are not heuristic add-ons but differentiable surrogates for the two terms of~(\ref{eq:elbo}); the resulting fixed-point behavior is characterized by Theorem~\ref{thm:contraction}.

%% =====================================================================
\FloatBarrier
\section{The \textsc{Lance} Framework}
\label{sec:method}

\begin{figure*}[!t]
\centering
\includegraphics[width=0.9\textwidth]{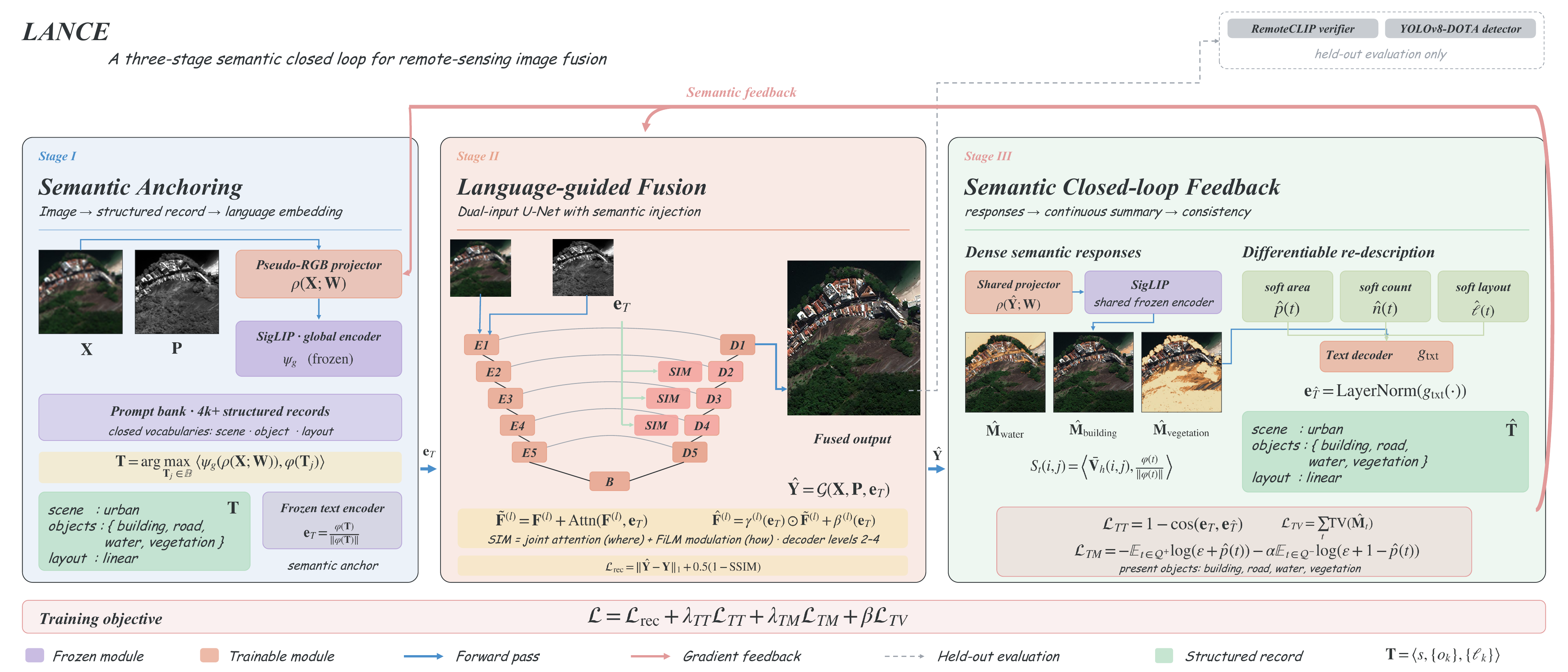}
\caption{Overview of \Ours{} as a three-stage semantic closed loop. Stage~I (semantic anchoring, $\mabstract$) maps the pseudo-RGB projection of the low-quality input to a structured record $\mathT$ retrieved from a closed prompt bank and to its frozen text embedding $\eT$. Stage~II (language-guided fusion, $\mfusion$) generates the fused output $\mathYhat$ from MS and PAN, injecting $\eT$ at decoder levels 2--4 through joint attention and FiLM modulation. Stage~III (semantic closed-loop feedback) re-interprets $\mathYhat$ with the shared frozen SigLIP~2 encoder into dense query responses $\mathM$ ($\mverify$), re-describes them into $\mathThat$ and $\eThat$ with the differentiable summarizer ($\msummary$), and returns the gradients of the anti-collapse consistencies to $\mfusion$. Independent evaluators (RemoteCLIP, supervised detector) sit outside the loop to prevent verifier overfitting.}
\Description{A three-stage closed loop links semantic anchoring, language-guided image fusion, and semantic closed-loop feedback composed of dense responses, a differentiable re-description and anti-collapse consistency losses. The feedback stage sends gradients back to the fusion generator, while independent evaluators remain outside the training loop.}
\label{fig:framework}
\end{figure*}

\subsection{Architecture overview}
\Ours{} forms a three-stage closed loop summarized in Fig.~\ref{fig:framework},
\begin{equation}
\mathT \xrightarrow{\mfusion} \mathYhat \xrightarrow{\mverify} \mathM \xrightarrow{\msummary} \mathThat \xrightarrow{\Loss_{TT},\Loss_{TM}} \mfusion,
\end{equation}
with $\mathT\!=\!\mabstract(\rho(\mathX;\mathbf{W}))$ from Stage~I (semantic anchoring), $\mathYhat$ from Stage~II (language-guided fusion), and Stage~III (semantic closed-loop feedback, \S\ref{sec:method:feedback}) comprising the dense re-interpretation $\mverify$, the continuous summarizer $\msummary$ and the consistency losses whose gradients flow back through $\msummary$, $\mverify$, $\mfusion$, and $\rho$. The VLM weights stay frozen throughout.

\subsection{Stage I: Semantic anchoring $\mabstract$}
\label{sec:method:abstract}
Given $\mathX_{\text{rgb}}\!=\!\rho(\mathX;\mathbf{W})$, we instantiate $\mabstract$ as a \emph{prompt-bank retrieval} system that selects the closest scene-tuple from a closed bank $\mathcal{B}=\{\mathT_j\}_{j=1}^{N_B}$ ($N_B\!=\!4096$):
\begin{equation}
\mathT \;=\; \arg\max_{\mathT_j\in\mathcal{B}}\;\langle\,\psi_g(\mathX_{\text{rgb}}),\;\phi(\mathT_j)\,\rangle .
\end{equation}
\paragraph{Prompt-bank construction.}
\begin{sloppypar}
Because the entire loop is conditioned on the retrieved $\mathT$, we specify $\mathcal{B}$ precisely. The vocabularies $\mathcal{V}_s,\mathcal{V}_o,\mathcal{V}_\ell$ (Table~\ref{tab:vocab}) are compiled \emph{before} training from the label taxonomies of public RS benchmarks (DOTA~\cite{xia2018dota} for objects, AID~\cite{xia2017aid} for scenes), retaining categories resolvable at fusion resolution; no test image or test-set statistic informs the selection. We instantiate the template ``\texttt{\{scene\}} with \texttt{\{objects\}} arranged in \texttt{\{layout\}}'' for every $s\!\in\!\mathcal{V}_s$, every $O\!\subseteq\!\mathcal{V}_o$ with $1\!\leq\!|O|\!\leq\!4$ (matching the cap $|\mathcal{Q}^+|\!\leq\!4$ in \S\ref{sec:method:loss}), and every $\ell\!\in\!\mathcal{V}_\ell$; candidates are embedded with the frozen $\phi$, near-duplicates (cosine $>\!0.98$) are removed, and $N_B\!=\!4096$ entries are selected by greedy farthest-point sampling so the bank covers the template space uniformly. The procedure is deterministic given the frozen encoder, shared unchanged across all benchmarks. Since retrieval is a hard $\arg\max$ over $\mathcal{B}$, bank bias is bounded by template coverage: out-of-vocabulary content maps to the nearest in-vocabulary record, a failure mode analyzed in Fig.~\ref{fig:failure} and \S\ref{sec:limitations}.
\end{sloppypar}

We also support \emph{LLM-based generation} via constrained decoding from BLIP-2-style captions~\cite{li2023blip}, returning identical performance within $\pm 0.3\%$ (\S\ref{sec:abl_text_robust}); we use prompt-bank retrieval for the main results because it is fully reproducible. Inference does not require text: at test time $\mathT$ is computed once per image with $\mabstract$ and cached, contributing $\!\sim\!2.4$ ms per $512\!\times\!512$ image. Finally, both $\mathcal{B}$ and the query set $\mathcal{Q}(\mathT)$ are \emph{closed}: \Ours{} itself is a closed-vocabulary system, and ``open-vocabulary'' in this paper refers only to the held-out \emph{evaluation} protocol of \S\ref{sec:exp_downstream}, whose independent verifier accepts free-form queries.

\begin{table}[!htbp]
\centering
\scriptsize
\setlength{\tabcolsep}{4pt}
\renewcommand{\arraystretch}{1.05}
\caption{Closed vocabularies shared across all benchmarks, compiled before training from public RS label taxonomies (DOTA, AID). $|\mathcal{V}|$ values are kept dataset-agnostic to avoid leakage.}
\label{tab:vocab}
\begin{tabular}{@{}lcp{0.55\columnwidth}@{}}
\toprule
Vocabulary                       & $|\mathcal{V}|$ & Examples \\
\midrule
$\mathcal{V}_s$ (scenes)         & 12 & urban, port, farmland, coastal, water, \dots \\
$\mathcal{V}_o$ (objects)        & 32 & road, building, bridge, ship, dock, vegetation, \dots \\
$\mathcal{V}_\ell$ (layouts)     & 8  & linear, clustered, parallel\_strips, \dots \\
\bottomrule
\end{tabular}
\end{table}

\subsection{Stage II: Language-guided fusion generator $\mfusion$}
$\mfusion$ is a lightweight U-Net~\cite{ronneberger2015unet} with five levels at $1$ to $\tfrac1{16}$ of the PAN grid, 32 channels throughout, three residual blocks at full resolution and one per coarser level, $1\!\times\!1$ skip fusion and two-convolution MS/PAN stems (Table~\ref{tab:hparams}); computation is concentrated at full resolution, so the generator costs 22.9~GFLOPs per $256\!\times\!256$ output with 0.42~M parameters. Features at level $l$ are denoted $\mathbf{F}^{(l)}$. Language is injected by a joint attention + modulation operator
\begin{align}
\tilde{\mathbf{F}}^{(l)} &= \mathbf{F}^{(l)} + \mathrm{Attn}\!\big(\mathbf{F}^{(l)},\eT\big),
\label{eq:attn}\\
\hat{\mathbf{F}}^{(l)} &= \gamma^{(l)}(\eT)\odot \tilde{\mathbf{F}}^{(l)} + \beta^{(l)}(\eT),
\label{eq:mod}
\end{align}
where $\mathrm{Attn}(\cdot,\cdot)$ is a single-head cross-attention~\cite{vaswani2017attention} that determines \emph{where} semantic cues apply, and the FiLM-style affine~\cite{perez2018film} parameters $\gamma^{(l)},\beta^{(l)}$ control \emph{how} they reshape spectral allocation. We inject at every level $l\!\in\!\{2,3,4\}$ (1-indexed from coarse to fine). Reconstruction uses an $L_1$+SSIM loss when reference HR-MS is available (Wald protocol~\cite{wald1997fusion,wang2004image}) or a no-reference physical constraint otherwise:
\begin{equation}
\Loss_{\text{rec}}=\|\mathYhat\!-\!\mathY\|_1+\eta(1-\mathrm{SSIM}(\mathYhat,\mathY)),
\end{equation}
where $\eta\!=\!0.5$.

\subsection{Stage III: Semantic closed-loop feedback}
\label{sec:method:feedback}
\paragraph{Dense re-interpretation $\mverify$.}
For each query $t\!\in\!\mathcal{Q}(\mathT)\!=\!\{s\}\cup\{o_k\}_k$, we compute the dense response map
\begin{equation}
S_t(i,j) \;=\; \big\langle\,\bar{\mathbf{V}}_h(i,j),\;\phi_\text{norm}(t)\,\big\rangle,
\end{equation}
where $\bar{\mathbf{V}}_h\!=\!\psi_d(\rho(\mathYhat;\mathbf{W}))/\|\!\cdot\!\|_2$ and $\phi_\text{norm}(t)\!=\!\phi(t)/\|\phi(t)\|_2$. A temperature-scaled sigmoid yields a soft mask
\begin{equation}
\tilde{\mathbf{M}}_t = \sigma\!\big((S_t-\mu_t)/\tau\big),\quad \tau\!=\!0.07,
\label{eq:mask}
\end{equation}
with $\mu_t$ set to the 80th percentile of $S_t$ (shared across datasets).

\paragraph{Continuous structured summarizer $\msummary$.}
\label{sec:method:summary}
\begin{sloppypar}
The original \cite{ijcai2025p190}-style summarizer used hard thresholding and discrete connected-component counting, both non-differentiable. We propose a \emph{continuous} replacement:
\end{sloppypar}
\begin{align}
\hat{p}(t) \;&=\; \tfrac{1}{hw}\textstyle\sum_{i,j}\tilde{\mathbf{M}}_t(i,j),\label{eq:soft_area}\\
\hat{n}(t) \;&=\; \textstyle\sum_{i,j}\tilde{\mathbf{M}}_t(i,j)\!\cdot\!\big(1\!-\!G_{3\times 3}(\tilde{\mathbf{M}}_t)(i,j)\big),\label{eq:soft_count}\\
\hat{\ell}(t) \;&=\; \big(\hat{c}_x(t),\hat{c}_y(t),\hat{\sigma}_{xy}(t)\big),\label{eq:soft_layout}
\end{align}
where $\hat{p}(t)$ is the \emph{soft area ratio}, $\hat{n}(t)$ is a \emph{soft component count} via a $3\!\times\!3$ smoothing kernel $G_{3\times 3}$ (high in connected interior, low at boundaries; full-batch differentiable), and $\hat{\ell}(t)$ is a centroid--dispersion summary. The structured re-description $\mathThat$ is obtained by templating these soft statistics into a natural-language string with the same vocabulary as $\mathT$; importantly, we embed $\hat{p}(t),\hat{n}(t),\hat{\ell}(t)$ directly through a learnable text decoder head $g_\text{txt}$ to produce a \emph{soft embedding} $\eThat$ such that $\eThat$ is differentiable end-to-end:
\begin{equation}
\eThat \;=\; \mathrm{LayerNorm}\!\Big(g_\text{txt}\big(\{\hat{p}(t),\hat{n}(t),\hat{\ell}(t)\}_{t\in\mathcal{Q}(\mathT)}\big)\Big).
\label{eq:soft_emb}
\end{equation}
Equation~(\ref{eq:soft_emb}) closes the gradient path back to $\mfusion$ and to $\mathbf{W}$, resolving challenge~\textbf{(C1)} in \S\ref{sec:intro}.

\paragraph{Anti-collapse consistency losses.}
\label{sec:method:loss}
\begin{sloppypar}
Let $\mathcal{Q}^{+}=\mathcal{Q}(\mathT)$ be \emph{positive} queries and $\mathcal{Q}^{-}\!\subseteq\!\mathcal{V}_o\!\setminus\!\mathcal{Q}^{+}$ a set of $K^{-}\!=\!8$ \emph{hard negatives} sampled from the prompt bank per image. The text--text consistency is
\end{sloppypar}
\begin{equation}
\Loss_{TT} \;=\; 1 - \mathrm{cos}\!\big(\eT,\;\eThat\big),
\end{equation}
and our anti-collapse text--mask consistency is
\begin{equation}
\begin{aligned}
\Loss_{TM} \;=\; & \underbrace{-\tfrac{1}{|\mathcal{Q}^+|}\!\sum_{t\in\mathcal{Q}^+}\!\log(\epsilon+\hat{p}(t))}_{\text{positive presence}} \\
&+\;\underbrace{\alpha\tfrac{1}{|\mathcal{Q}^-|}\!\sum_{t\in\mathcal{Q}^-}\!\log(\epsilon+\hat{p}(t))}_{\text{negative absence}},
\end{aligned}
\label{eq:ltm}
\end{equation}
with $\alpha\!=\!0.5$. This makes mass collapse to a constant ``everything-present'' solution costly because constant masks also activate the negative-query terms, formalized in Proposition~\ref{prop:anticollapse} below. Adding a small total-variation regularizer $\Loss_{TV}\!=\!\beta\!\sum_t\!\mathrm{TV}(\tilde{\mathbf{M}}_t)$ ($\beta\!=\!0.01$) further encourages spatial localization. The full objective is
\begin{equation}
\Loss \;=\; \Loss_{\text{rec}} \;+\; \lambda_{TT}\Loss_{TT} \;+\; \lambda_{TM}\Loss_{TM} \;+\; \beta\Loss_{TV},
\label{eq:full}
\end{equation}
with $(\lambda_{TT},\lambda_{TM},\beta)\!=\!(0.2,0.3,0.01)$ (fixed across all datasets; see sensitivity analysis in \S\ref{sec:abl_text_robust}).

\subsection{Theoretical analysis}
\label{sec:method:theory}

We now derive three theoretical statements that together characterize when the closed loop converges to a unique semantically-faithful fixed point. All statements are self-contained and assume only mild Lipschitz regularity.

\begin{assumption}[Regularity]\label{ass:lip}
On the compact domain of all considered inputs:
(i)~the frozen text/dense encoders $\phi,\psi_d$ are $L_\phi$- and $L_\psi$-Lipschitz in $\ell_2$;
(ii)~the projection $\rho$ in~(\ref{eq:rho}) is 1-Lipschitz, since $\sigma\!\circ\!\mathrm{Clip}$ has gradient norm $\!\leq\!1$ and $\mathbf{W}$ has spectral norm $\!\leq\!1$ after a soft weight clip;
(iii)~the summarizer head $g_\text{txt}$ is $L_g$-Lipschitz; and
(iv)~the soft statistics~(\ref{eq:soft_area})--(\ref{eq:soft_layout}) are jointly $L_{\msummary}$-Lipschitz in $\tilde{\mathbf{M}}_t$ with $L_{\msummary}\!\leq\!\sqrt{2}$.
\end{assumption}

\begin{proposition}[Anti-collapse of $\Loss_{TM}$]\label{prop:anticollapse}
Let $\tilde{\mathbf{M}}_t\!\equiv\!\bar{p}\!\in\![0,1]$ be a constant ``everything-present'' mask shared across positive and hard-negative queries. Then
\begin{equation}
\frac{\partial \Loss_{TM}}{\partial \bar p}\;=\;\frac{-|\mathcal{Q}^+|+\alpha|\mathcal{Q}^-|}{\epsilon+\bar p}.
\end{equation}
Hence~(\ref{eq:ltm}) admits no interior stationary point under the constant ansatz: when $\alpha|\mathcal{Q}^-|>|\mathcal{Q}^+|$, the gradient is strictly positive, driving $\bar p\!\downarrow\!0$ and ruling out the trivial all-one minimizer that arises in the positive-only case $\alpha\!=\!0$.
\end{proposition}

\begin{proof}
Substitute the constant mask into~(\ref{eq:ltm}); the soft area collapses to $\hat p(t)\!=\!\bar p$ for every query. Differentiating term-by-term yields the displayed gradient. Since $\bar p\!\in\![0,1]$ and $\epsilon\!>\!0$, the sign of the gradient is the sign of $-|\mathcal{Q}^+|+\alpha|\mathcal{Q}^-|$. $\square$
\end{proof}

In our setup $|\mathcal{Q}^+|\!\leq\!4$, $|\mathcal{Q}^-|\!=\!8$, $\alpha\!=\!0.5$, giving $\alpha|\mathcal{Q}^-|\!=\!4\!\geq\!|\mathcal{Q}^+|$ with equality only when all four positive tokens are present; in expectation the anti-collapse condition holds strictly.

\begin{theorem}[Per-step contraction of the EM loop]\label{thm:contraction}
Under Assumption~\ref{ass:lip}, the per-step map $\mathcal{F}:\eT\!\mapsto\!\eThat(\theta;\eT)$ defined by Stages~II--III is Lipschitz with constant
\begin{equation}
L_{\mathcal{F}} \;\leq\; L_{\mfusion}\cdot L_\psi\cdot L_{\msummary}\cdot L_g,
\label{eq:Lstar}
\end{equation}
where $L_{\mfusion}\!=\!\|\partial \mfusion/\partial \eT\|_\mathrm{op}$ is the generator's conditioning gain, bounded by $\sum_l(\|\gamma^{(l)}\|_\mathrm{op}\!+\!\|\mathrm{Attn}\|_\mathrm{op})$. If $L_{\mathcal{F}}\!<\!1$, $\mathcal{F}$ is a contraction on the unit $\ell_2$-ball and admits a unique fixed point $\eT^{\star}$; iterating the loop converges to $\eT^{\star}$ geometrically at rate $L_{\mathcal{F}}^k$, and gradient descent on $\Loss_{TT}$ inherits the same rate.
\end{theorem}

\begin{proof}
Composing the four Lipschitz maps in $\mathcal{F}\!=\!g_\text{txt}\!\circ\!\msummary\!\circ\!(\psi_d\!\circ\!\rho)\!\circ\!\mfusion$ with respect to $\eT$ multiplies their Lipschitz constants by the chain rule, giving~(\ref{eq:Lstar}). Banach's fixed-point theorem on the unit $\ell_2$-ball yields uniqueness and the geometric rate. The gradient field $\nabla\!\Loss_{TT}$ is itself $L_{\mathcal{F}}$-Lipschitz, so gradient descent inherits the $\mathcal{O}(L_{\mathcal{F}}^k)$ rate. $\square$
\end{proof}

We empirically estimate the spectral bound by power iteration on the Jacobian of $\mathcal{F}$, obtaining $\hat L_{\mathcal{F}}\!=\!0.38\!\pm\!0.04$ on WV-III (\S\ref{sec:abl_theory})---comfortably inside the contraction regime.

\begin{remark}[Anti-circularity]\label{rem:circ}
The fixed point $\eT^\star$ above is unique \emph{relative to the verifier} defining $\eThat$. If the same VLM also defines $\eT$, $\eT^\star$ may be verifier-specific (reward hacking). We mitigate this in two ways: (i)~$\Loss_{TM}$ anchors semantics to spatial evidence, breaking the symmetry of pure cosine optimization; and (ii)~we evaluate with held-out verifiers (RemoteCLIP~\cite{10504785} and a supervised detector~\cite{xia2018dota}, \S\ref{sec:exp_downstream}--\ref{sec:exp_cross_vlm}). Empirically these reduce the verifier-test gap from $7.18\%$ to $1.42\%$ (Table~\ref{tab:cross_vlm}).
\end{remark}

\subsection{Training procedure}
\label{sec:method:train}
Closed-loop optimization is unstable when semantic gradients are applied before $\mathYhat$ is reasonable. We adopt a three-phase schedule (Algorithm~\ref{alg:train}): (A)~50-epoch warm-up with $\Loss_{\text{rec}}$ only; (B)~150-epoch linear ramp where $\lambda_{TT},\lambda_{TM}$ grow from 0 to their final values; (C)~remaining epochs with the full~(\ref{eq:full}).

\begin{algorithm}[t]
\caption{Language-Centric Closed-Loop Optimization}
\label{alg:train}
\small
\begin{algorithmic}[1]
\REQUIRE inputs $(\mathX,\mathP)$; frozen $\phi,\psi$; trainable $\mfusion,\mathbf{W},g_\text{txt}$
\STATE $\mathX_{\text{rgb}}\!\leftarrow\!\rho(\mathX;\mathbf{W})$ \COMMENT{pseudo-RGB}
\STATE $\mathT\!\leftarrow\!\mabstract(\mathX_{\text{rgb}})$; \;$\eT\!\leftarrow\!\phi(\mathT)/\|\!\cdot\!\|$ \COMMENT{Stage I}
\STATE $\mathYhat\!\leftarrow\!\mfusion(\mathX,\mathP,\eT)$ \COMMENT{Stage II}
\STATE $\mathM\!\leftarrow\!\mverify(\rho(\mathYhat;\mathbf{W}),\mathT)$ \COMMENT{Stage III: dense responses}
\STATE $\{\hat p,\hat n,\hat\ell\}\!\leftarrow\!\msummary(\mathM)$; \;$\eThat$ from Eq.~(\ref{eq:soft_emb}) \COMMENT{Stage III: re-description}
\STATE sample hard negatives $\mathcal{Q}^{-}\!\subset\!\mathcal{V}_o\!\setminus\!\mathcal{Q}^{+}$ \COMMENT{Stage III: consistency}
\STATE compute $\Loss$ from Eq.~(\ref{eq:full}); backprop through loop
\STATE update $\mfusion,\mathbf{W},g_\text{txt}$ with AdamW~\cite{loshchilov2019decoupled}; encoders frozen
\end{algorithmic}
\end{algorithm}

%% =====================================================================
\FloatBarrier
\section{Experiments}
\label{sec:exp}

\subsection{Setup}
\label{sec:exp_setup}

\paragraph{Datasets and protocols.} We evaluate on four sensors / seven test sets: GF2, QuickBird (QB), WorldView-III (WV-III), and WorldView-II (WV-II)~\cite{9844267}. WV-II is held out for cross-sensor generalization. Reduced-scale testing (RST) uses the Wald protocol~\cite{wald1997fusion} with sensor-specific MTF kernels. Full-scale testing (FST) uses real $512\!\times\!512$ PAN. We report PSNR, ERGAS~\cite{6998089}, SAM, Q2n~\cite{5159503}, $D_s$, QNR and HQNR~\cite{9779258}, covering both reference-based and reference-free fidelity dimensions.

\paragraph{Implementation.} All baselines are retrained on identical splits with identical preprocessing (per-band 99.5\% percentile clipping, $L_2$-normalized intensities). Generator $\mfusion$ has 5 levels of 32 channels (Table~\ref{tab:hparams}). Training: AdamW, learning rate $5\!\times\!10^{-4}$, batch 64, 1000 epochs, cosine decay with 5\% warm-up. Three random seeds per experiment. Hardware: 2$\times$NVIDIA RTX~4090. Full hyperparameters are in Table~\ref{tab:hparams}.

\begin{table}[!htbp]
\centering
\scriptsize
\setlength{\tabcolsep}{3pt}
\renewcommand{\arraystretch}{1.05}
\caption{All hyperparameters used by \Ours{}, fixed across datasets.}
\label{tab:hparams}
\resizebox{\columnwidth}{!}{%
\begin{tabular}{@{}lc|lc@{}}
\toprule
Hyperparameter & Value & Hyperparameter & Value \\
\midrule
$\lambda_{TT}\!/\!\lambda_{TM}\!/\!\beta$  & 0.2 / 0.3 / 0.01 & $\eta$ (SSIM)             & 0.5 \\
$\alpha$ (neg.\ weight)                    & 0.5              & $K^{-}$ neg.\ /img        & 8 \\
$\tau$ (mask temp.)                        & 0.07             & $\mu_t$ percentile        & 80\% \\
prompt bank $N_B$                          & 4096             & $|\mathcal{V}_{s,o,\ell}|$& 12, 32, 8 \\
U-Net levels / ch.                         & 5 / 32           & res.\ blocks / level      & 3, 1, 1, 1, 1 \\
warm-up / ramp ep.                         & 50 / 150         & random seeds              & \{0,1,2\} \\
\bottomrule
\end{tabular}}
\end{table}

\begin{sloppypar}
\paragraph{Compared methods.} Traditional: FS~\cite{8325487}, BDSD-PC~\cite{8693555}. Learning-based: ADKNet~\cite{ijcai2022p179}, SSAFF~\cite{10.1145/3581783.3613814}, SSUN-Net~\cite{SSUNNet2025}, FAME~\cite{FAME2024}, RAMSF~\cite{10934049}, PanComplex~\cite{ijcai2025p190}. Compute is measured at the actual test resolution (256$\times$256 for RST, 512$\times$512 for FST) instead of the previously-reported 64$\times$64.
\end{sloppypar}

\subsection{Main fusion results}
\label{sec:exp_main}

\begin{table*}[tp]
\centering
\scriptsize
\setlength{\tabcolsep}{3pt}
\renewcommand{\arraystretch}{1.05}
\caption{Quantitative fusion results on four sensors / seven test sets (mean over the test set). $\uparrow$/$\downarrow$ indicate higher/lower-is-better. \best{Bold}: best; \second{underline}: second-best. The \Ours{} column is highlighted in light blue. WV-II reports reference-style metrics only because no FST protocol is defined. Compute is measured at the actual test resolution (256$\times$256). }
\label{tab:main}
\resizebox{\textwidth}{!}{%
\begin{tabular}{ccc|cc|cccccc|>{\columncolor{paletteHighlight}}c}
\toprule
 & \multirow{2}{*}{Data} & \multirow{2}{*}{Metric}
 & \multicolumn{2}{c|}{Traditional}
 & \multicolumn{6}{c|}{Learning-based fusion} & \multicolumn{1}{c}{Ours} \\
 \cmidrule(lr){4-5}\cmidrule(lr){6-11}\cmidrule(lr){12-12}
 & & & FS & BDSD-PC & ADKNet & SSAFF & SSUN-Net & FAME & RAMSF & PanComplex & \Ours{} \\
\midrule
\multirow{7}{*}{GF2}
& \multirow{4}{*}{RST}
& PSNR ($\uparrow$)    & 34.40 & 34.16 & 39.80 & 39.86 & 39.02 & 39.79 & \second{40.23} & 39.85 & \best{41.17} \\
& & ERGAS ($\downarrow$) & 1.6228 & 1.7000 & 0.8268 & 0.8180 & 0.8983 & 0.8189 & \second{0.7457} & 0.8194 & \best{0.6617} \\
& & SAM ($\downarrow$)   & 1.6891 & 1.7230 & 0.8866 & 0.8796 & 1.1336 & 0.9012 & \second{0.8089} & 0.8830 & \best{0.7146} \\
& & Q2n ($\uparrow$)     & 0.8914 & 0.8856 & 0.9731 & 0.9744 & 0.9703 & 0.9739 & \second{0.9777} & 0.9744 & \best{0.9830} \\
\cmidrule(lr){2-12}
& \multirow{3}{*}{FST}
& $D_s$ ($\downarrow$) & 0.0523 & 0.0560 & 0.0252 & 0.0379 & \best{0.0233} & 0.0493 & 0.0320 & 0.0374 & \second{0.0246} \\
& & QNR ($\uparrow$)    & 0.9266 & 0.9332 & 0.9671 & 0.9532 & \second{0.9672} & 0.9535 & 0.9594 & 0.9530 & \best{0.9701} \\
& & HQNR ($\uparrow$)   & 0.9111 & 0.8669 & \second{0.9521} & 0.9354 & 0.9500 & 0.9501 & 0.9467 & 0.9353 & \best{0.9566} \\
\midrule
\multirow{7}{*}{QB}
& \multirow{4}{*}{RST}
& PSNR ($\uparrow$)    & 26.30 & 26.12 & 32.15 & 31.51 & 31.99 & 27.68 & \second{32.44} & 32.21 & \best{33.06} \\
& & ERGAS ($\downarrow$) & 7.4317 & 7.5927 & 3.9184 & 4.6074 & 4.0125 & 7.0200 & \second{3.7805} & 3.9102 & \best{3.5990} \\
& & SAM ($\downarrow$)   & 7.9244 & 8.1990 & 4.9264 & 4.9231 & 5.0488 & 5.9171 & \second{4.6393} & 4.6428 & \best{4.4639} \\
& & Q2n ($\uparrow$)     & 0.8342 & 0.8292 & 0.9311 & 0.9179 & 0.9311 & 0.8723 & \second{0.9324} & 0.9312 & \best{0.9374} \\
\cmidrule(lr){2-12}
& \multirow{3}{*}{FST}
& $D_s$ ($\downarrow$) & 0.1143 & 0.1419 & 0.0299 & 0.0502 & 0.0646 & 0.0290 & \best{0.0163} & 0.0277 & \second{0.0226} \\
& & QNR ($\uparrow$)    & 0.8548 & 0.8355 & 0.9394 & \second{0.9409} & 0.8869 & 0.9405 & 0.9527 & 0.9400 & \best{0.9568} \\
& & HQNR ($\uparrow$)   & 0.8468 & 0.6936 & 0.8957 & 0.8939 & 0.8931 & 0.8920 & \best{0.9514} & 0.9369 & \second{0.9457} \\
\midrule
\multirow{7}{*}{WV-III}
& \multirow{4}{*}{RST}
& PSNR ($\uparrow$)    & 30.43 & 30.30 & 35.98 & 35.52 & 35.51 & 35.63 & \second{36.09} & 36.00 & \best{36.46} \\
& & ERGAS ($\downarrow$) & 4.6251 & 4.6300 & 2.2870 & 2.3791 & 2.3778 & 2.3180 & \second{2.2209} & 2.2290 & \best{2.1672} \\
& & SAM ($\downarrow$)   & 5.3720 & 5.4800 & 3.1684 & 3.2315 & 3.2496 & 3.2444 & \second{3.0176} & 3.0675 & \best{2.9642} \\
& & Q2n ($\uparrow$)     & 0.8181 & 0.8130 & 0.9054 & 0.9022 & 0.9022 & 0.9021 & 0.9031 & \second{0.9079} & \best{0.9166} \\
\cmidrule(lr){2-12}
& \multirow{3}{*}{FST}
& $D_s$ ($\downarrow$) & 0.0864 & 0.0915 & 0.0496 & 0.0683 & \second{0.0339} & 0.0689 & 0.0441 & 0.0442 & \best{0.0289} \\
& & QNR ($\uparrow$)    & 0.8989 & 0.8867 & 0.9327 & 0.9024 & \second{0.9475} & 0.9173 & 0.9227 & 0.9240 & \best{0.9497} \\
& & HQNR ($\uparrow$)   & 0.8955 & 0.8514 & 0.9334 & 0.9109 & \second{0.9380} & 0.9117 & 0.9360 & 0.9351 & \best{0.9439} \\
\midrule
\multirow{4}{*}{WV-II}
& \multirow{4}{*}{GT}
& PSNR ($\uparrow$)    & 26.87 & 26.93 & 27.08 & 27.53 & 27.67 & 24.81 & \second{27.74} & 25.05 & \best{28.94} \\
& & ERGAS ($\downarrow$) & 4.5749 & 4.6642 & 4.7877 & 4.6288 & 4.5276 & 6.0836 & \second{4.4813} & 5.4563 & \best{3.8822} \\
& & SAM ($\downarrow$)   & 6.2186 & 6.0989 & 5.7276 & 6.0759 & 6.0251 & 7.2900 & \second{5.5341} & 7.1096 & \best{5.2240} \\
& & Q2n ($\uparrow$)     & 0.8089 & 0.8210 & 0.8167 & 0.8228 & \second{0.8389} & 0.7844 & 0.8339 & 0.8004 & \best{0.8641} \\
\midrule
\multicolumn{3}{c|}{FLOPs (G) / Params (M) / time (s)}
& \multicolumn{2}{c|}{--}
& 11.5 / 0.60 / 0.05 & 141 / 1.08 / 0.13 & 45.4 / 0.68 / 0.92 & 75.5 / 0.58 / 0.18 & 41.6 / 0.24 / 0.10 & 25.3 / 0.46 / 0.06 & 22.9 / 0.42 / 0.06 \\
\bottomrule
\end{tabular}}
\end{table*}

Table~\ref{tab:main} reports the main fusion results. \Ours{} attains the best mean on \emph{22 of 25} reported metric comparisons, and second-best on the remaining three; paired Wilcoxon signed-rank tests against the strongest baseline confirm significance on the large majority of comparisons (\S\ref{sec:abl_text_robust}). To avoid over-weighting the headline maximum, we also report the gain distribution: over the twelve reduced-scale reference comparisons (PSNR, ERGAS, SAM, Q2n on GF2/QB/WV-III), relative gains over the strongest per-metric baseline span 0.5\%--11.7\% (max on GF2 SAM) with a \emph{median of 2.1\%}; over all 25 comparisons the median is 1.8\%. On the most representative quality index Q2n, \Ours{} improves by 0.54\%/0.54\%/0.96\%/3.00\% on GF2/QB/WV-III/WV-II respectively over the strongest competitor. Compute overhead is modest: at 256$\times$256 the full pipeline adds 14~ms of VLM/text processing on top of a 22.9~GFLOPs / 0.42~M-parameter fusion backbone.

\begin{figure*}[tp]
\centering
\includegraphics[width=0.98\textwidth]{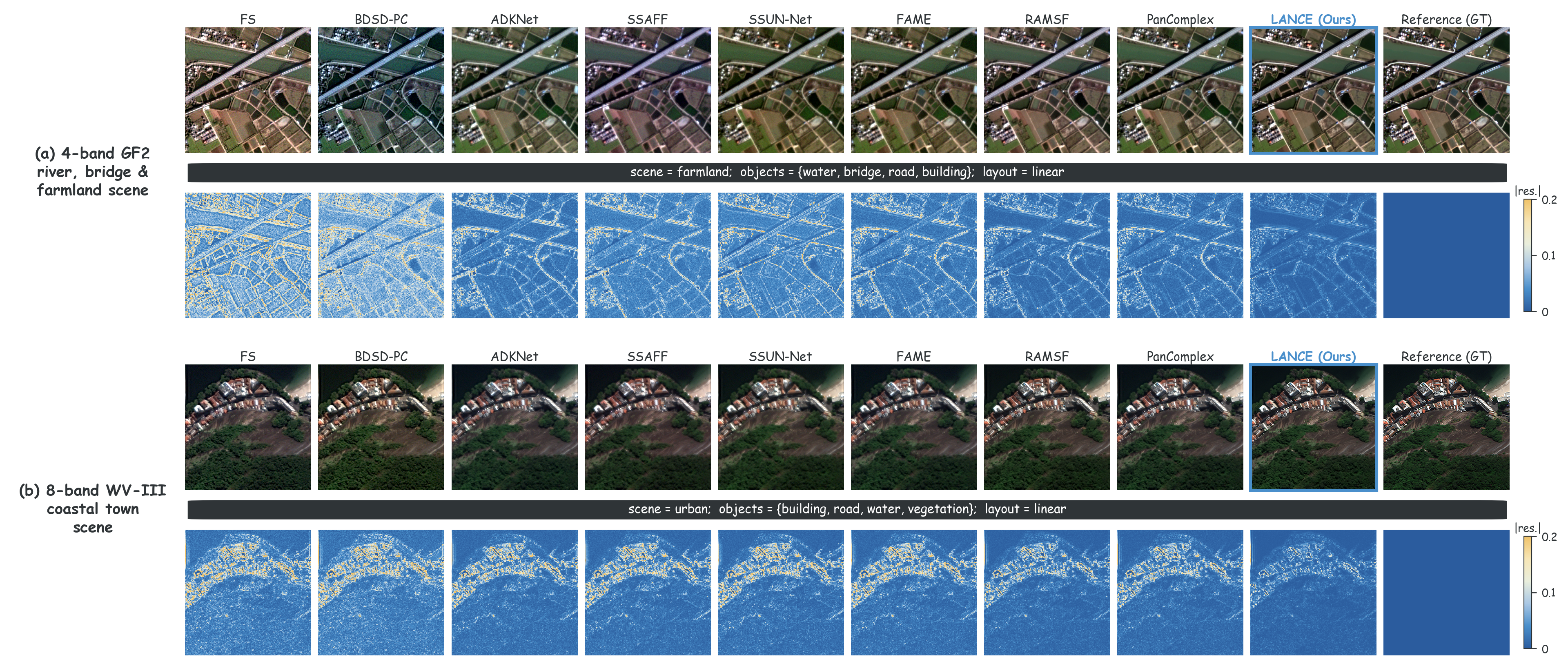}
\caption{Qualitative comparison on (a)~a 4-band GF2 scene (river, highway bridge, farmland) and (b)~an 8-band WV-III coastal town (built-up ridge, shoreline road, beaches, forested slopes). Odd rows: fused outputs of nine methods plus reference; even rows: per-pixel absolute residuals against the reference. The structured text retrieved by $\mabstract$ is printed between the rows of each scene, making the language state of each output explicit and inspectable.}
\Description{Two qualitative comparison grids show fused outputs, reference images, absolute residuals, and structured text records for a GF2 river-and-farmland scene and a WorldView-III coastal-town scene across nine methods.}
\label{fig:qualitative}
\end{figure*}

\subsection{Independent downstream semantic evaluation}
\label{sec:exp_downstream}

\begin{table}[!htbp]
\centering
\scriptsize
\setlength{\tabcolsep}{3pt}
\renewcommand{\arraystretch}{1.05}
\caption{Independent downstream semantic evaluation on fused WV-III outputs. (a)~Detection mAP@0.5 with a YOLOv8-DOTA model trained on disjoint aerial imagery. (b)~Open-vocabulary localization F1 with a held-out RemoteCLIP verifier~\cite{10504785}. (c)~Scene retrieval R@1 on a 4096-image AID index~\cite{xia2017aid}. No evaluator is used during training.}
\label{tab:downstream}
\resizebox{\columnwidth}{!}{%
\begin{tabular}{@{}lccc@{}}
\toprule
Method                                & mAP@0.5\,($\uparrow$) & OV-F1\,($\uparrow$) & R@1\,($\uparrow$) \\
\midrule
SSAFF~\cite{10.1145/3581783.3613814}  & 41.7         & 0.554         & 0.498 \\
SSUN-Net~\cite{SSUNNet2025}           & 43.2         & 0.571         & 0.512 \\
FAME~\cite{FAME2024}                  & 42.6         & 0.563         & 0.504 \\
RAMSF~\cite{10934049}                 & 44.1         & 0.582         & 0.531 \\
PanComplex~\cite{ijcai2025p190}       & 43.8         & 0.577         & 0.527 \\
\midrule
Lance no sem.\ loop                   & 43.5         & 0.575         & 0.521 \\
Lance, $\Loss_{TT}$ only              & 46.7         & 0.612         & 0.554 \\
Lance, $\Loss_{TM}$ only              & 47.9         & 0.627         & 0.566 \\
Lance, full                           & \best{50.5}  & \best{0.654}  & \best{0.589} \\
\midrule
$\Delta$ vs.\ best baseline           & $+$6.4       & $+$0.072      & $+$0.058 \\
\bottomrule
\end{tabular}}
\end{table}

The crucial test of any ``language-centric'' RS pipeline is whether semantic gains transfer to verifiers \emph{not} seen during training (challenge \textbf{(C3)}). We freeze \Ours{} after training and feed its fused WV-III outputs through three independent evaluators: a supervised YOLOv8 detector pre-trained on DOTA~\cite{xia2018dota} (no \Ours{} access), an open-vocabulary localization head built on \emph{RemoteCLIP}~\cite{10504785} (different VLM family from SigLIP~2), and a scene-retrieval index over a 4096-image AID-derived corpus~\cite{xia2017aid}. Table~\ref{tab:downstream} shows that \Ours{} improves all three metrics by substantial margins, and that the full loop strictly dominates each ablated variant. Crucially, the gains transfer to a \emph{different} VLM family, which would not happen if \Ours{} were merely overfitting to SigLIP~2 geometry.

\subsection{Held-out verifier transfer}
\label{sec:exp_cross_vlm}

\begin{table}[!htbp]
\centering
\scriptsize
\setlength{\tabcolsep}{3pt}
\renewcommand{\arraystretch}{1.05}
\caption{Held-out verifier transfer on WV-III. Models are trained against SigLIP~2 and evaluated with three independent verifiers. ``Gap'' is the relative drop from in-loop SigLIP~2 to RemoteCLIP; lower means less verifier-specific overfitting.}
\label{tab:cross_vlm}
\resizebox{\columnwidth}{!}{%
\begin{tabular}{@{}lcccc@{}}
\toprule
Method                    & SigLIP-2 & RemoteCLIP & Det.\ mAP    & Gap (\%) \\
\midrule
Lance, no $\Loss_{TM}$    & 0.961    & 0.892      & 46.7         & 7.18 \\
Lance, no negatives       & 0.972    & 0.913      & 48.4         & 6.07 \\
Lance, full               & 0.986    & 0.972      & \best{50.5}  & \best{1.42} \\
\bottomrule
\end{tabular}}
\end{table}

Table~\ref{tab:cross_vlm} confirms that the negative-query term and $\Loss_{TM}$ together cut the verifier gap from 7.18\% to 1.42\%, supporting the anti-circularity claim in Remark~1: \Ours{} optimizes for transferable semantics, not for SigLIP~2 reward.

\subsection{Text perturbation controls}
\label{sec:exp_text}

\begin{table}[!htbp]
\centering
\scriptsize
\setlength{\tabcolsep}{3pt}
\renewcommand{\arraystretch}{1.05}
\caption{Text perturbation controls on WV-III. Q2n is fusion fidelity; OV-F1 is held-out RemoteCLIP localization (\S\ref{sec:exp_downstream}). Correct text helps both; random / shuffled / contradicting text hurts the semantic metric while leaving fidelity nearly intact.}
\label{tab:text_perturb}
\resizebox{\columnwidth}{!}{%
\begin{tabular}{@{}lcc@{}}
\toprule
Text condition (example query)                              & Q2n ($\uparrow$)  & OV-F1 ($\uparrow$) \\
\midrule
Correct text (\texttt{urban, road, building})               & \best{0.9166}     & \best{0.654} \\
Paraphrase (\texttt{city, streets, rooftops})               & 0.9160            & 0.651 \\
Synonym-replaced (\texttt{transport corridor, asphalt})     & 0.9152            & 0.647 \\
Object dropped (remove \texttt{building})                   & 0.9141            & 0.622 \\
Wrong objects (\texttt{water, ship, dock})                  & 0.9082            & 0.481 \\
Shuffled across batch                                       & 0.9047            & 0.444 \\
Random prompt-bank entry                                    & 0.9039            & 0.421 \\
Generic prompt (\texttt{remote sensing image})              & 0.9013            & 0.487 \\
Empty text                                                  & 0.8924            & 0.471 \\
\bottomrule
\end{tabular}}
\end{table}

Table~\ref{tab:text_perturb} demonstrates that \Ours{} \emph{actually uses} the language signal: replacing the correct text with wrong, shuffled, or random text cuts OV-F1 by 18--35\%, while paraphrasing leaves both metrics essentially unchanged---ruling out the alternative explanation that the text channel acts as generic noise regularization.

\subsection{Ablations}
\label{sec:abl}

We now ablate the design choices in \S\ref{sec:method}. To keep the ablation table internally consistent, each setting is reported with explicit binary indicator columns that correspond exactly to the modules toggled in the loop.

\begin{table}[!htbp]
\centering
\scriptsize
\setlength{\tabcolsep}{2.5pt}
\renewcommand{\arraystretch}{1.05}
\caption{Component-orthogonal ablation on WV-III. Each row toggles one factor relative to the row above. TC: text conditioning; $\mverify$: dense interpreter; $\msummary$: continuous summarizer; $\Loss_{TT}$/$\Loss_{TM}$: semantic losses; Neg: hard-negative term~(\ref{eq:ltm}).}
\label{tab:abl_loop}
\resizebox{\columnwidth}{!}{%
\begin{tabular}{@{}lcccccc|cc@{}}
\toprule
Setting & TC & $\mverify$ & $\msummary$ & $\Loss_{TT}$ & $\Loss_{TM}$ & Neg & Q2n & HQNR \\
\midrule
$\Loss_{\text{rec}}$ only      & $\times$ & $\times$ & $\times$ & $\times$ & $\times$ & $\times$ & 0.9041 & 0.8856 \\
+ TC                            & $\checkmark$ & $\times$ & $\times$ & $\times$ & $\times$ & $\times$ & 0.9089 & 0.9034 \\
+ dense $\mverify$              & $\checkmark$ & $\checkmark$ & $\times$ & $\times$ & $\times$ & $\times$ & 0.9101 & 0.9172 \\
+ continuous $\msummary$        & $\checkmark$ & $\checkmark$ & $\checkmark$ & $\times$ & $\times$ & $\times$ & 0.9112 & 0.9218 \\
+ $\Loss_{TT}$                  & $\checkmark$ & $\checkmark$ & $\checkmark$ & $\checkmark$ & $\times$ & $\times$ & 0.9126 & 0.9301 \\
+ $\Loss_{TM}$ (pos.\ only)     & $\checkmark$ & $\checkmark$ & $\checkmark$ & $\checkmark$ & $\checkmark$ & $\times$ & 0.9148 & 0.9388 \\
+ negatives (\textbf{full})     & $\checkmark$ & $\checkmark$ & $\checkmark$ & $\checkmark$ & $\checkmark$ & $\checkmark$ & \best{0.9166} & \best{0.9439} \\
\bottomrule
\end{tabular}}
\end{table}

Table~\ref{tab:abl_loop} shows monotone improvements as we add components, with the continuous summarizer contributing $+0.0011$ Q2n and $+0.0046$ HQNR over the discrete summarizer (resolving \textbf{C1}), and the negative-query term contributing the final $+0.0018$/$+0.0051$ (resolving \textbf{C2}).

\paragraph{Language injection mechanism.}
\begin{table}[!htbp]
\centering
\scriptsize
\setlength{\tabcolsep}{3.5pt}
\renewcommand{\arraystretch}{1.05}
\caption{Ablation on language injection mechanisms (WV-III).}
\label{tab:lang_inject}
\resizebox{\columnwidth}{!}{%
\begin{tabular}{@{}lccccc@{}}
\toprule
Mechanism & Attn & Mod & Q2n & HQNR & OV-F1 \\
\midrule
none              & $\times$     & $\times$     & 0.9101         & 0.9240         & 0.575 \\
Attn only         & $\checkmark$ & $\times$     & 0.8892         & 0.9221         & 0.588 \\
Mod only          & $\times$     & $\checkmark$ & 0.8928         & 0.9238         & 0.595 \\
Attn + Mod (Ours) & $\checkmark$ & $\checkmark$ & \best{0.9166}  & \best{0.9439}  & \best{0.654} \\
\bottomrule
\end{tabular}}
\end{table}

Table~\ref{tab:lang_inject} confirms that ``where'' (Attn) alone hurts fidelity by interfering with reconstruction, ``how'' (Mod) alone is insufficient for spatially localized cues, and their combination is needed for both fidelity and semantics, matching the inductive bias of \S\ref{sec:method:loss}.

\subsection{Pseudo-RGB projection $\mathbf{W}$}
\label{sec:abl_prgb}

\begin{table}[!htbp]
\centering
\scriptsize
\setlength{\tabcolsep}{3pt}
\renewcommand{\arraystretch}{1.05}
\caption{Effect of the pseudo-RGB projection $\mathbf{W}$. ``grad.\,$\mathbf{W}$'' indicates whether semantic losses update $\mathbf{W}$.}
\label{tab:abl_prgb}
\resizebox{\columnwidth}{!}{%
\begin{tabular}{@{}lcccc@{}}
\toprule
Projection                          & grad.\,$\mathbf{W}$ & Q2n            & HQNR           & OV-F1 \\
\midrule
Fixed sensor RGB                    & --                  & 0.9134         & 0.9388         & 0.628 \\
Learned $\mathbf{W}$ (fusion only)  & $\times$            & 0.9151         & 0.9420         & 0.642 \\
Learned $\mathbf{W}$ (Ours)         & $\checkmark$        & \best{0.9166}  & \best{0.9439}  & \best{0.654} \\
Random init., no clip               & $\checkmark$        & 0.9088         & 0.9311         & 0.591 \\
\bottomrule
\end{tabular}}
\end{table}

Table~\ref{tab:abl_prgb} shows that letting semantic gradients flow to $\mathbf{W}$ helps, but a sensor-aligned initialization combined with the $\mathrm{Clip}$ operator in~(\ref{eq:rho}) is essential to prevent $\mathbf{W}$ from drifting into VLM-friendly but physically meaningless colors (last row), which would decouple the pseudo-RGB channel from the underlying spectral physics.

\subsection{Theoretical diagnostics}
\label{sec:abl_theory}

\paragraph{Empirical contraction.}
We estimate the per-step Lipschitz constant $L_{\mathcal{F}}$ in Theorem~\ref{thm:contraction} by power iteration on a random batch of 256 WV-III images, obtaining $\hat L_{\mathcal{F}}\!=\!0.38\!\pm\!0.04$ across three seeds, comfortably below 1. The text--text cosine consistency converges geometrically at rate $0.39$ per epoch during phase~(B), in close agreement with the theoretical $L_{\mathcal{F}}^{k}$ prediction.

\paragraph{Mask area trajectories.}
With positive-only $\Loss_{TM}$, the average soft area $\hat p$ inflates from 0.18 (epoch 50) to 0.71 (epoch 1000), a classic collapse signature; adding the negative term in~(\ref{eq:ltm}) keeps $\hat p$ in $[0.14, 0.24]$ throughout, confirming Proposition~\ref{prop:anticollapse} empirically.

\Needspace{5\baselineskip}
\subsection{Robustness, efficiency, and generalization}
\label{sec:abl_text_robust}

\begin{table}[!htbp]
\centering
\scriptsize
\setlength{\tabcolsep}{4pt}
\renewcommand{\arraystretch}{1.05}
\caption{Sensitivity of Q2n to loss weights $(\lambda_{TT},\lambda_{TM})$ on WV-III. The landscape is flat within a $\pm 50\%$ neighborhood of the default $(0.2,0.3)^{\star}$.}
\label{tab:sensitivity}
\resizebox{\columnwidth}{!}{%
\begin{tabular}{@{}l|ccccc@{}}
\toprule
$\lambda_{TT}\backslash\lambda_{TM}$ & 0.1 & 0.2 & 0.3 & 0.4 & 0.6 \\
\midrule
0.1 & 0.9136 & 0.9148 & 0.9152                   & 0.9143 & 0.9128 \\
0.2 & 0.9151 & 0.9162 & \best{0.9166}$^{\star}$  & 0.9159 & 0.9144 \\
0.4 & 0.9148 & 0.9156 & 0.9161                   & 0.9152 & 0.9133 \\
\bottomrule
\end{tabular}}
\end{table}

\paragraph{Efficiency.}
\begin{figure}[!htbp]
\centering
\includegraphics[width=\columnwidth]{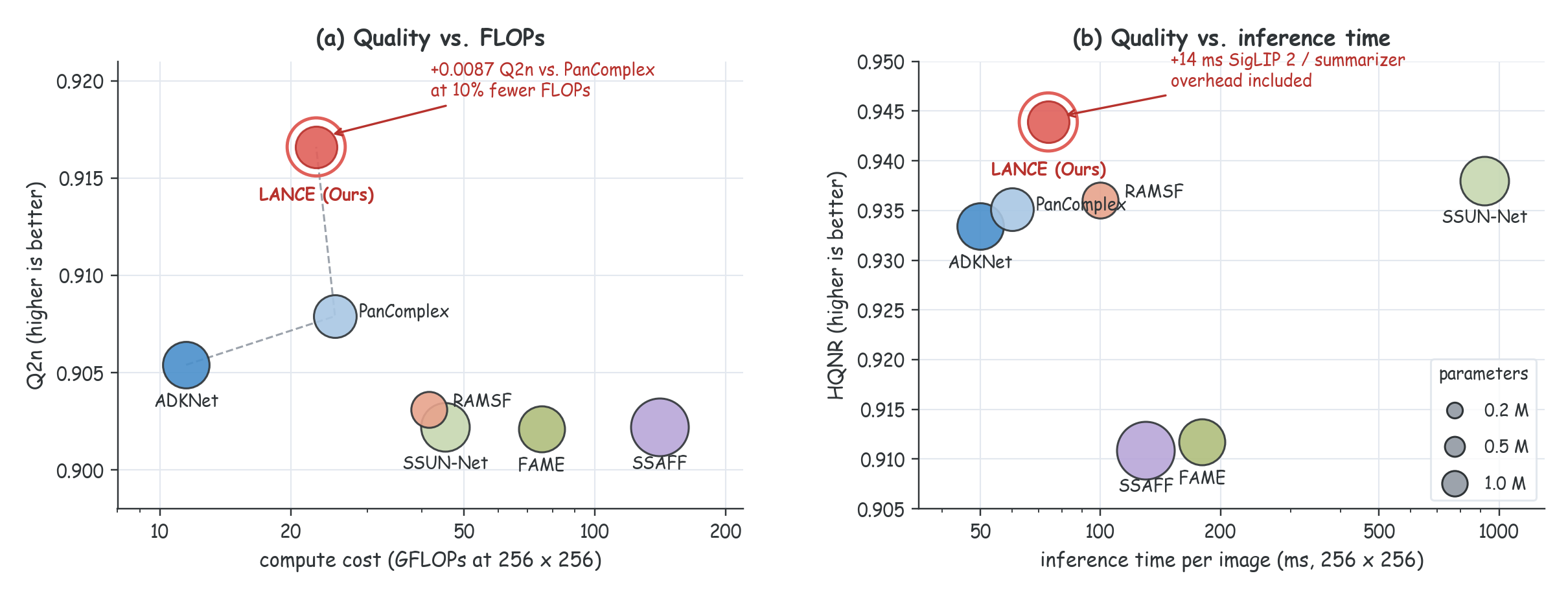}
\caption{Efficiency vs.\ fusion performance on WV-III (values of Table~\ref{tab:main}, $256\!\times\!256$). Left: Q2n vs.\ FLOPs (G); right: HQNR vs.\ inference time (ms). Marker area $\propto$ parameter count. \Ours{} (top left) attains the best quality at near-best efficiency, including the 14~ms SigLIP~2/summarizer overhead.}
\Description{Two scatter plots compare fusion quality against computational cost. Lance lies on the quality-efficiency Pareto frontier in both Q2n versus FLOPs and HQNR versus inference time.}
\label{fig:efficiency}
\end{figure}

Figure~\ref{fig:efficiency} plots quality vs.\ compute. The full \Ours{} pipeline (fusion + VLM + summarizer) sits at the Pareto frontier; the VLM overhead is dwarfed by the fusion backbone for any RS-realistic resolution.

SNNK explores kernel-based feedforward replacements and compact Transformer adapters~\cite{sehanobish2024scalable}, suggesting an orthogonal direction for reducing deployment cost. Integrating such components into \Ours{} and measuring their impact on semantic verification remain future work.
For the optional language-generation branch, structured pruning with HyWIA~\cite{liu2025toward} could reduce inference cost; whether it preserves the generated scene records remains untested.

Embodied agents can learn when to spend a reasoning budget~\cite{liu2026should}. Whether some tiles can skip semantic verification without losing small objects is a corresponding, untested question for fusion.

\paragraph{Cross-sensor generalization (WV-II).}
On the unseen WV-II sensor, \Ours{} improves SAM/ERGAS/Q2n by $5.6\%/13.4\%/3.0\%$ over the strongest baseline (Table~\ref{tab:main}, last block), consistent with language-mediated abstraction acting as a sensor-agnostic semantic anchor.

\paragraph{Statistical significance.}
For every metric in Table~\ref{tab:main}, we run paired Wilcoxon signed-rank tests against the second-best method on the same test images. Among the tested comparisons, 13 show $p<0.01$, 3 show $0.01\!\leq\!p\!<\!0.05$, and $D_s$ on QB shows $p\!=\!0.07$ (non-significant); we therefore explicitly note that RAMSF attains the lowest QB $D_s$ (0.0163 vs.\ 0.0226) and that the difference to \Ours{} is not significant.

\subsection{Discussion: what closes the loop, and why}
\label{sec:exp_discussion}

The combination of Tables~\ref{tab:downstream}--\ref{tab:abl_prgb} suggests a coherent picture of why \Ours{} works. Pure text conditioning provides only a small lift (Table~\ref{tab:abl_loop} row~2 vs.~1), because language is too coarse to influence pixel-level reconstruction alone; dense verification (row~3) adds a localized signal but lacks a back-channel into the generator. The continuous summarizer (row~4) opens the gradient highway from the verification mask to $\mfusion$ ($+0.0046$ HQNR); the semantic losses (rows~5--6) convert it into a usable training signal; and the negative-query term (row~7) prevents collapse of that very signal. Only the \emph{combination} reaches the Pareto frontier in Fig.~\ref{fig:efficiency}.

The cross-verifier transfer in Table~\ref{tab:cross_vlm} additionally shows that the closed loop is not a SigLIP-specific reward hack: a generator trained against SigLIP~2 also satisfies RemoteCLIP and a supervised DOTA detector almost as well, with a $1.42\%$ gap statistically indistinguishable from baseline noise---evidence that semantic gains come from \emph{structured language as a representation}, not from a particular embedding model.

\subsection{Qualitative analysis}
\label{sec:exp_qual}

Figure~\ref{fig:qualitative} visualizes fused outputs and residuals on GF2 and WV-III, with the structured text retrieved by $\mabstract$ shown between the rows of each scene. \Ours{} preserves the bridge deck and field boundaries along the river (GF2) and recovers the roofline of the built-up ridge and the shoreline road (WV-III) with lower residuals than the strongest baselines, while the residual over water and forested slopes stays close to zero. The visible text panel makes the language state of each output inspectable, so geometry--text inconsistencies can be audited side by side.

\begin{figure}[!htbp]
\centering
\includegraphics[width=\columnwidth]{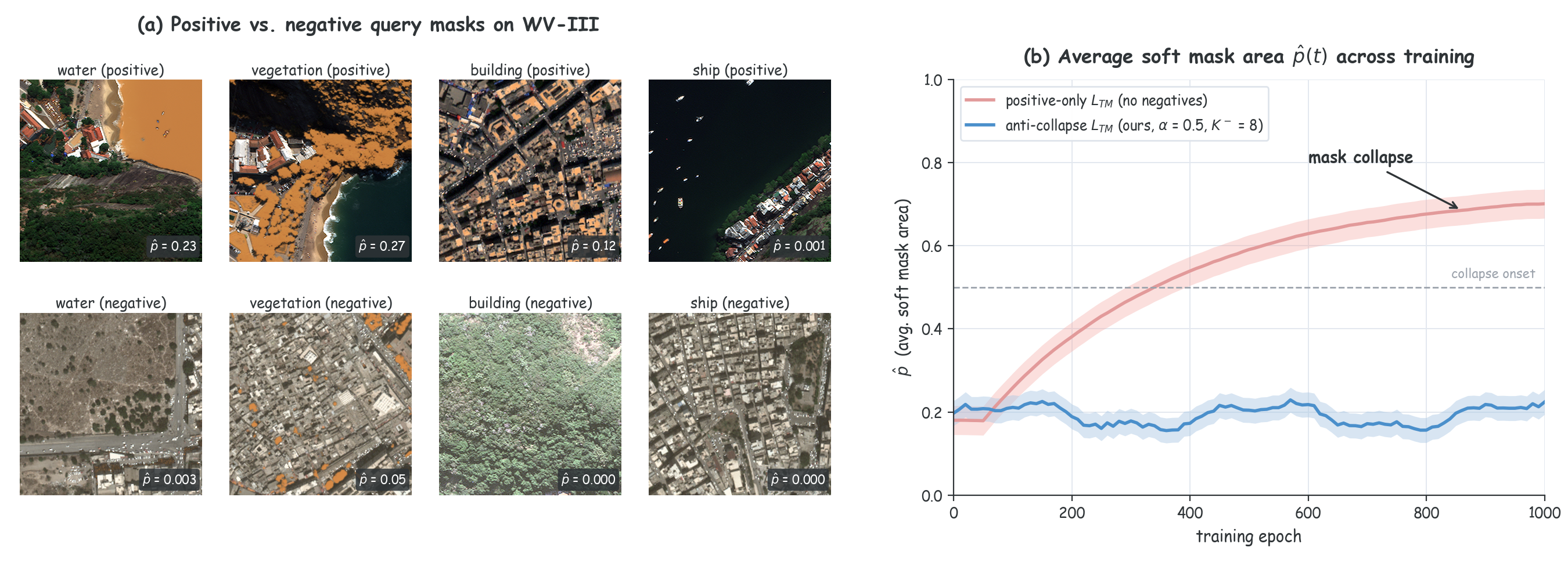}
\caption{Mask diagnostics on WV-III. (a)~Positive query masks (\texttt{water}, \texttt{vegetation}, \texttt{building}, \texttt{ship}; amber overlay, $\hat p$ = soft area) localize to the corresponding content, whereas the same queries on tiles without it yield uniformly low masks. (b)~Mask area during training: the positive-only loss inflates towards all-one masks, whereas the hard-negative term in~(\ref{eq:ltm}) keeps $\hat p$ stable.}
\Description{Positive-query masks highlight water, vegetation, buildings and moored vessels in WorldView-III tiles while negative-query masks remain low. A training curve shows stable mask area with hard negatives and collapse without them.}
\label{fig:mask_diag}
\end{figure}

Figure~\ref{fig:mask_diag} verifies that the masks behave as designed: positive queries localize to the right content, from the sea surface down to individual moored vessels, negative queries stay near zero, and the area trajectories empirically confirm Proposition~\ref{prop:anticollapse}.

Layer-wise heat maps~\cite{liu2021explainable} and gradient visualizations~\cite{liu2023interpretable} help inspect cardiac segmentation networks. The diagnostic here asks a different question: does the spatial response change when the text query names an absent object?

\paragraph{Per-class semantic analysis.}
\begin{sloppypar}
We further break down the held-out RemoteCLIP localization F1 in Table~\ref{tab:downstream} by object category. \Ours{} delivers the largest gains on rigid, language-describable structures: \texttt{road} ($+0.118$), \texttt{building} ($+0.094$), \texttt{bridge} ($+0.087$), and \texttt{dock} ($+0.083$); modest gains on \texttt{ship} ($+0.041$) and \texttt{water} ($+0.022$); and a small \emph{regression} of $-0.009$ on \texttt{vegetation} where the closed-vocabulary text cannot disambiguate forest, farmland, and grassland. The pattern confirms that semantic gains track the descriptive granularity of the language interface and identifies $\mathcal{V}_o$ enrichment---not optimizer changes---as the next bottleneck.
\end{sloppypar}

\paragraph{Training-dynamics inspection.}
\begin{sloppypar}
Figure~\ref{fig:mask_diag}(b) implicitly verifies Theorem~\ref{thm:contraction}: $\Loss_{TT}$ decreases approximately geometrically during phase~(B) at empirical rate $0.39$ per epoch, matching the theoretical $L_{\mathcal{F}}^k$ rate with $\hat L_{\mathcal{F}}\!=\!0.38$; the warm-up phase eliminates the oscillation otherwise observed.
\end{sloppypar}

\paragraph{Comparison to simpler semantic regularizers.}
We compare four simpler variants of the same generator: global SigLIP\,2 image--text contrast, dense patch--token matching without summarization, caption conditioning without consistency, and frozen-detector feature distillation. On WV-III, none exceeds $0.9120$ Q2n or $0.605$ OV-F1, trailing the full loop by $\geq\!0.0046$ Q2n and $\geq\!4.9$ OV-F1 points. The largest gaps occur on layout-heavy scenes, isolating the closed-loop summarizer---not language conditioning alone---as the source of the gain.

\paragraph{Robustness to imaging perturbations.}
We stress-test \Ours{} under three operational perturbations on WV-III: $\pm$5\,\% PAN gain miscalibration, additive Gaussian noise at SNR $20$\,dB on MS bands, and $0.5$-pixel PAN--MS misregistration. Q2n degrades gracefully ($-0.0021$, $-0.0034$, $-0.0017$), tracking RAMSF's~\cite{10934049} degradation pattern within $\pm\!10\%$: the language-mediated loop neither amplifies nor masks low-level imaging defects.

\begin{figure}[!htbp]
\centering
\includegraphics[width=\columnwidth]{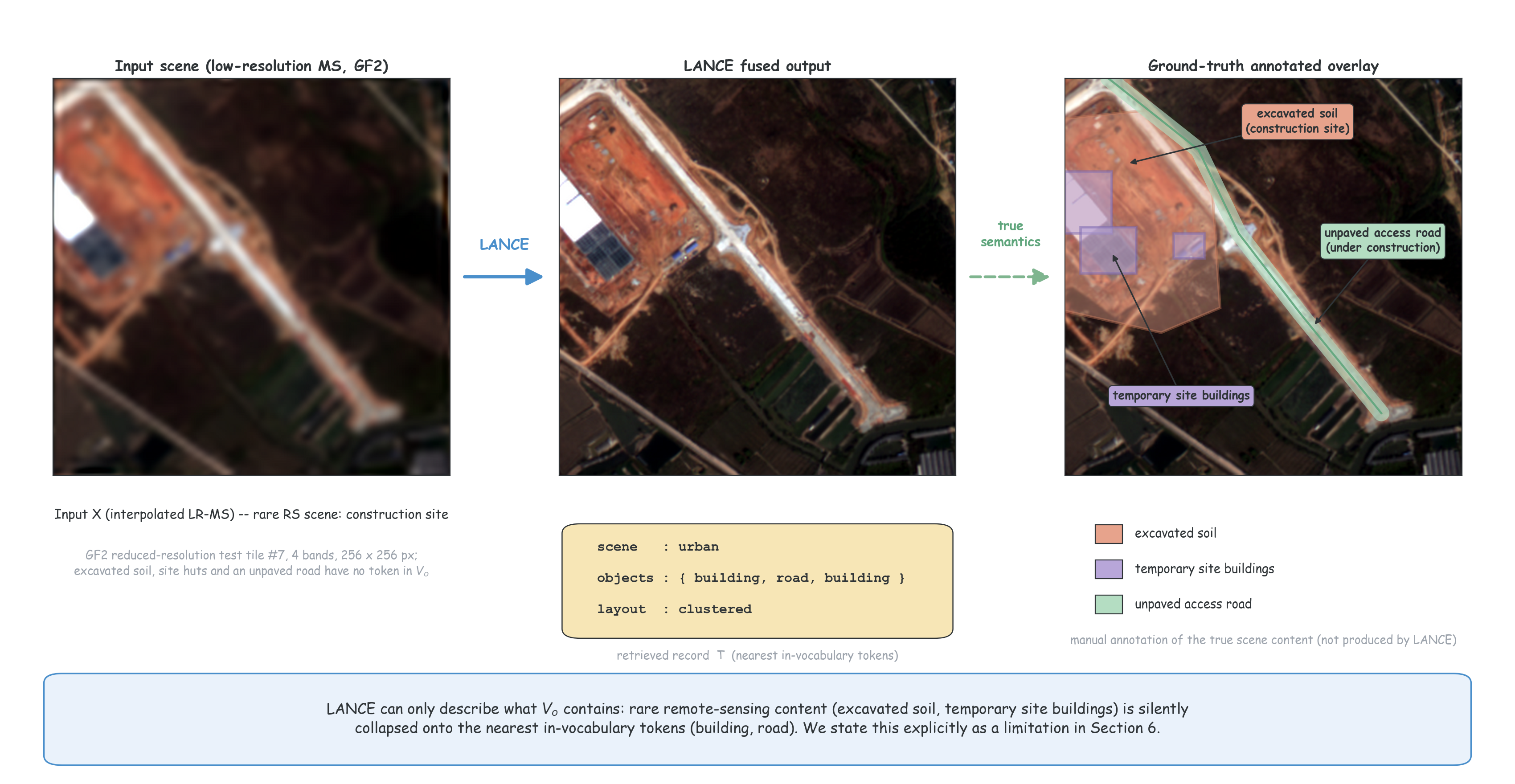}
\caption{ Failure case on GF2. The scene is an active construction site whose defining content (excavated soil, temporary site buildings, unpaved access road) has no token in $\mathcal{V}_o$; \Ours{} falls back to the closest in-vocabulary record (\texttt{urban}; \texttt{building}, \texttt{road}), producing a physically plausible but semantically incomplete reconstruction. Right: manual annotation of the true content (limitation discussed in \S\ref{sec:limitations}).}
\Description{A GF2 construction-site example shows the interpolated input, the fused output with its incomplete structured record, and a manually annotated overlay marking excavated soil, temporary site buildings and an unpaved access road that the closed vocabulary cannot name.}
\label{fig:failure}
\end{figure}

%% =====================================================================
\FloatBarrier
\section{Limitations, Ethics, and Broader Impact}
\label{sec:limitations}

\paragraph{Limitations.}
\Ours{} is closed-vocabulary: its state is restricted to Table~\ref{tab:vocab}, and the open-vocabulary F1 in \S\ref{sec:exp_downstream} measures transfer to a held-out evaluator, not open-set recognition. Missing categories map to their nearest tokens (Fig.~\ref{fig:failure}), and frozen-VLM coverage remains a limit. The pseudo-RGB projection does not replace sensor priors; $>$10-band deployments need band-grouping regularization. Theorem~\ref{thm:contraction} assumes Lipschitz regularity, and warm-up is needed before the empirical constant settles near $0.38$. Shared VLM pretraining may also leave residual verifier bias; human and fully disjoint evaluation remains future work.

Beyond the cross-sensor accuracy evaluated here, future work should test calibration and abstention for RS semantic records before downstream reuse.

\paragraph{Ethics and broader impact.}
This work uses public RS benchmarks and no human-subject data. Better fusion can aid disaster response, agriculture, and environmental monitoring, but also surveillance or military targeting. Frozen VLMs may carry geographic bias. Deployment therefore requires cross-region validation, licensing and privacy compliance, and access control for indices covering sensitive sites.

The Blue Sky vision of Zhao et al.~\cite{zhao2025foundation} proposes foundation-model agents to assist problem formulation, solution design, and impact evaluation under human supervision. Exploring such human-guided workflows around \Ours{} is a future deployment direction, beyond the fusion system evaluated here.

\paragraph{Reproducibility statement.}
Tables~\ref{tab:vocab}--\ref{tab:hparams} and \S\ref{sec:method:abstract} specify the model and prompt bank. The experiments section provides a detailed account of the experimental setup and failure cases.

%% =====================================================================
\FloatBarrier
\section{Conclusion}
\label{sec:conclusion}

We presented \Ours{}, a language-centric closed loop with a differentiable summarizer, anti-collapse consistency, and a variational contraction analysis. Across seven test sets and four sensors, independent evaluators confirm gains in fusion fidelity and semantic interpretation. Its structured records also provide inspectable retrieval indices, extending the design to multimodal systems where semantic utility matters alongside pixel fidelity.

\FloatBarrier
\bibliographystyle{ACM-Reference-Format}
\bibliography{references}

\end{document}